%% file: iclr2019_conference.tex
\documentclass{article} 
\usepackage{times}
\usepackage{authblk}
\usepackage{microtype}
\usepackage{subcaption}
\usepackage{booktabs} 
\usepackage[utf8]{inputenc} 
\usepackage[T1]{fontenc}    
\usepackage{booktabs}       
\usepackage{amsfonts}       
\usepackage{nicefrac}       
\usepackage{microtype}      
\usepackage{float}
\usepackage{graphicx,subcaption}
\usepackage{paralist}
\usepackage{multirow}
\usepackage{amsmath,bbm,amssymb}
\usepackage{algorithm}
\usepackage[noend]{algorithmic}
\usepackage{amsthm}
\usepackage{tabularx}
\usepackage[round]{natbib}

\newcommand{\bx}{\mathbf{x}}
\newcommand{\by}{\mathbf{y}}
\newcommand{\bz}{\mathbf{z}}

\newcommand{\revise}{REVISE }

\input{math_commands.tex}

\input{notation.tex}

\usepackage{hyperref}
\usepackage{url}

\hypersetup{
    colorlinks,
    linkcolor={blue},
    citecolor={blue},
    urlcolor={blue}
}

\title{Towards Realistic Individual Recourse and Actionable Explanations in Black-Box Decision Making Systems}

\author[1]{Shalmali Joshi}
\affil[1]{Vector Institute, \texttt{shalmali@vectorinstitute.ai}}
\author[2]{Oluwasanmi Koyejo}
\author[2]{Warut Vijitbenjaronk}
\affil[2]{UIUC, \texttt{sanmi@illinois.edu, wdvijit@gmail.com}}
\author[3]{Been Kim}
\affil[3]{Google Brain, \texttt{beenkim@google.com}}
\author[4]{Joydeep Ghosh}
\affil[4]{UT Austin, \texttt{jghosh@utexas.edu}}

\begin{document}
\maketitle

\begin{abstract}
Machine learning based decision making systems are increasingly affecting humans. An individual can suffer an undesirable outcome under such decision making systems (e.g. denied credit) irrespective of whether the decision is fair or accurate. Individual recourse pertains to the problem of providing an actionable set of changes a person can undertake in order to improve their outcome. We propose a recourse algorithm that models the underlying data distribution or manifold. We then provide a mechanism to generate the smallest set of changes that will improve an individual's outcome. This mechanism can be easily used to provide recourse for any differentiable machine learning based decision making system. Further, the resulting algorithm is shown to be applicable to both supervised classification and causal decision making systems. Our work attempts to fill gaps in existing fairness literature that have primarily focused on discovering and/or algorithmically enforcing fairness constraints on decision making systems. This work also provides an alternative approach to generating counterfactual explanations.
\end{abstract}

\input{intro.tex}
\input{related_work.tex}
\input{model.tex}
\input{experiments.tex}
\input{conclusion.tex}

\bibliography{recourse}
\bibliographystyle{plainnat}

\input{appendix.tex}

\end{document}

%% file: math_commands.tex
\usepackage{amsmath,amsfonts,bm}

\def\eqref#1{equation~\ref{#1}}

\def\1{\bm{1}}

\DeclareMathAlphabet{\mathsfit}{\encodingdefault}{\sfdefault}{m}{sl}
\SetMathAlphabet{\mathsfit}{bold}{\encodingdefault}{\sfdefault}{bx}{n}

\DeclareMathOperator*{\argmin}{arg\,min}

%% file: notation.tex
\mathchardef\mhyphen="2D

\newcommand{\vertiii}[1]{{\left\vert\kern-0.25ex\left\vert\kern-0.25ex\left\vert #1
    \right\vert\kern-0.25ex\right\vert\kern-0.25ex\right\vert}}

\def\bm{{\mathbf{m}}}

\def\bx{{\mathbf{x}}}
\def\by{{\mathbf{y}}}
\def\bz{{\mathbf{z}}}

\def\bI{{\mathbf{I}}}

\def\bM{{\mathbf{M}}}



%% file: intro.tex
\section{Introduction}\label{sec:intro}
Machine learning (ML) algorithms have become widely deployed in domains that directly impact humans, including the criminal justice system~\citep{angwin2016machine}, clinical healthcare~\citep{Callahan2017279}, credit approval~\citep{siddiqi2012credit}, etc. As consumers of services, individuals might face unfavorable outcomes when subjected to such automated decision making and should almost always be provided with concrete mechanisms that would allow them to improve their outcomes. This set of "actionable changes" a consumer can resort to are called \emph{recourse}~\citep{ustun2018actionable}. This task of determining and providing changes to improve outcomes for a consumer is aligned with providing safeguards to individuals' rights but has received much less attention in the otherwise rich fairness literature.

This work attempts to fill this gap by proposing a new framework to provide algorithmic recourse. That is, in situations where an ML model provides an unsatisfactory outcome for an individual subjected to such automated decision making, we provide a list of attribute changes that will effectively help the individual to improve their outcome. The key to our approach is to characterize data manifold and/or distributions of the individuals who may be subjected to decision making using an ML based system to encourage actionable and realistic changes. We then provide an optimization framework to traverse the data manifold via its latent representation. This allows to sample the set of high probability paths of changes that are close to the original attributes, yet improve on their current (\emph{undesirable}) outcome.

The mechanism proposed here follows sample paths that are minimal and restricted along the data manifold toward improving outcomes. Our model avoids suggesting a recourse that is unlikely or unrealistic under the distribution of the client population. For instance, without restricting to the data manifold, a change in income might be the smallest change to improving client A's outcome but can be much less likely under their given circumstances. Such circumstantial difficulties (where clients similar to our hypothetical client A cannot realistically make drastic incomes changes) are implicitly codified by modeling the data distribution. The shortest path to improving client A's outcome along the manifold might simply be to increase monthly payments by a fixed amount. Thus these changes suggested by the proposed method can be considered to be more effective and achievable. 

This is one of the key ways in which our approach differs from existing literature on actionable recourse~\citep{ustun2018actionable}. Further,~\citet{ustun2018actionable} focus on linear ML models deployed in the algorithmic decision-making systems. Their mechanism may potentially provide a set of changes (called flipsets) that could affect other variables not included in the flipset. By characterizing the data manifold, such issues can be mitigated. Their work is nonetheless seminal in highlighting the problem as well as providing an easy to access toolkit for auditing recourses for white-box linear classifiers using integer programming. Our model additionally provides recourse across different class of models, like classification and causal models.

%% file: related_work.tex
\section{Related Work}\label{sec:related_work}
While work in providing recourse for individuals subject to algorithmic decision-making is extremely limited,\mbox{~\citet{ustun2018actionable}} have highlighted the issue as an important step toward mitigating algorithmic injustice. They allow recourses that optimize a user specified cost function and prevent recourses from changing immutable variables like age, sex, gender as is desirable using linear integer programming. Our contributions add to this work, and generates recourses for a much larger class of machine learning decision making systems, while also providing a competitive benchmark for generating counterfactual explanations and justifications for an outcome. Our approach characterizes the entire data distribution and provides actionable recourse by taking the shortest sample paths along the data manifold to improve outcomes. Thus the generated recourses need not be samples that present in training data as done by~\citep{wachter2017counterfactual} but are still realistically achievable without restricting to the class of linear decision making systems. Additionally our mechanism can be used to compare a wider class of models beyond supervised classification to provide such recourse. Finally, we also demonstrate how the presence of specific confounding in attributes affects recourse quality under the model as a means to provide counterfactual explanations.


Note that we do not view the proposed mechanism as a means to mitigate liability issues for the institution deploying algorithmic decision--making, which is the focus of providing counterfactual explanations~\citep{wachter2017counterfactual,dhurandhar2018explanations}. Counterfactual explanations are aimed at addressing why a model provided a specific outcome. For instance, if a credit loan is approved,~\citet{grath2018interpretable} show the margin (for each client attribute) by which the loan was approved and also provide a margin by which a loan is denied. To do so, they look at changes to client features in manhattan distance normalized by Mean Absolute Deviation to encourage sparsity of their explanations. However this does not ensure that the changes follow realistic sample paths along the data manifold and in some cases, the target profile is restricted to be an observed training sample. A side effect of such algorithms is that the resulting counterfactual may lie outside the ambient data domain.

%% file: model.tex
\section{A Framework for Individual Recourse}
First, we consider a supervised classification system trained with a differentiable loss function and model class. For credit loan approval, our goal would be to suggest a (minimal) set of changes to an individual's financial asset profile, in order to improve their outcome. Without loss of generality, we expose our model for a binary classification system. Let ${f}_{\phi} : \mathbb{R}^{d} \rightarrow \{-1, 1\}$ be the target classifier under which recourse is to be determined for $\bx \in \mathbb{R}^d$ sampled from the probability distribution $p(\mathbf{x})$. Let $y \in \{-1, 1\}$ be the set of outcomes where -1 is a undesirable outcome (denial of credit) and 1 is the desired outcome (approval of credit). We assume that the estimate is learned using the loss function $\ell (f_{\phi}(\bx), y)$ for the class of models $f$ parametrized by $\phi$ where $y \in \{-1, 1\}$ are the true labels. Let $c : \mathbb{R}^d \times \mathbb{R}^d \rightarrow \mathbb{R}_{+}$ be some distance measure in the sample space, $[d] = \{1, 2, \cdots, d\}$ and the $abs(.)$ be the absolute value function. Let $\mathbf{x}_i^*$ be the $i^{th}$ attribute of the data point $\mathbf{x}*$. Conceptually, for an individual instance $\mathbf{x}^*$ with outcome $\by^* \neq 1$ we would like to estimate $\mathbf{x}'$ such that:
\begin{align}\label{eq:classification_recourse}
\begin{split}
 \mathbf{x}' = & {\arg\min}_{\mathbf{x} | p(\mathbf{x}) > \gamma}  c(\mathbf{x}^*, \mathbf{x})\\
 & {\text{s. t. }} f_{\phi}(\mathbf{x}') = y' = 1
\end{split}
\end{align}
 where $\gamma >0$ parametrizes how likely the sample is under the distribution $p(\mathbf{x})$. The recourse for this individual can then be determined as the tuple $\{(d_i,\delta_{d_i} = \mathbf{x}_i^*- \mathbf{x}'_i) \, \forall i \in [d] \, \text{s. t. } abs(\delta_{d_i}) > 0 \}$. That is, recourse is the set of all attributes and corresponding changes that would improve the outcome for $\mathbf{x}^*$. While we jointly provide recourse over all attributes, the order in which a consumer may attempt to change these is not provided by the algorithm. Furthermore, while our exposition does not associate different weights to each attribute, this can be easily accommodated by using an appropriate distance function $c(.)$. In order to efficiently estimate such a change, we convert this problem to a constrained optimization framework by first characterizing the data distribution using a generative model. We now briefly describe the class of generative models used in this work.

\paragraph{Generative Models} can be described as stochastic procedures that generate samples $\bx \in \mathbb{R}^d$ from the data distribution $p(\bx)$. The two most significant types are the Variational Auto-Encoders (VAEs)~\citep{kingma2013auto} and Generative Adversarial Networks (GANs)~\citep{goodfellow2014generative}. Generative models generally assume that an underlying latent variable $\bz \in \mathbb{R}^k$ is mapped to the ambient data domain $\bx \in \mathbb{R}^d$ using a deterministic  function $\mathcal{G}_{\theta}$ parametrized by $\theta$, usually as a deep neural network. We skip further details in the interest of space. GANs employ an adversarial framework by using a discriminator that tries to classify generated samples from the original samples (rendering the probabilistic generative model to be implicit) and VAEs maximize an approximation to the data likelihood. The approximation obtained in a VAE has an encoder-decoder structure of conventional autoencoders~\citep{doersch2016tutorial}. One can obtain a latent representation of any data sample within the latent embedding using the trained encoder network of the VAE. While GANs do not train an associated encoder, recent advances in adversarially learned inference like BiGANs~\citep{dumoulin2016adversarially,donahue2016adversarial} can be utilized to obtain the latent embedding. We denote this encoder function (if trained and available) by $\mathcal{F}_{\psi} : \mathbb{R}^d \rightarrow \mathbb{R}^k$ (parametrized by $\psi$).

\begin{figure}[!h]
\centering
\begin{minipage}{0.35\textwidth}
\includegraphics[width=0.94\textwidth]{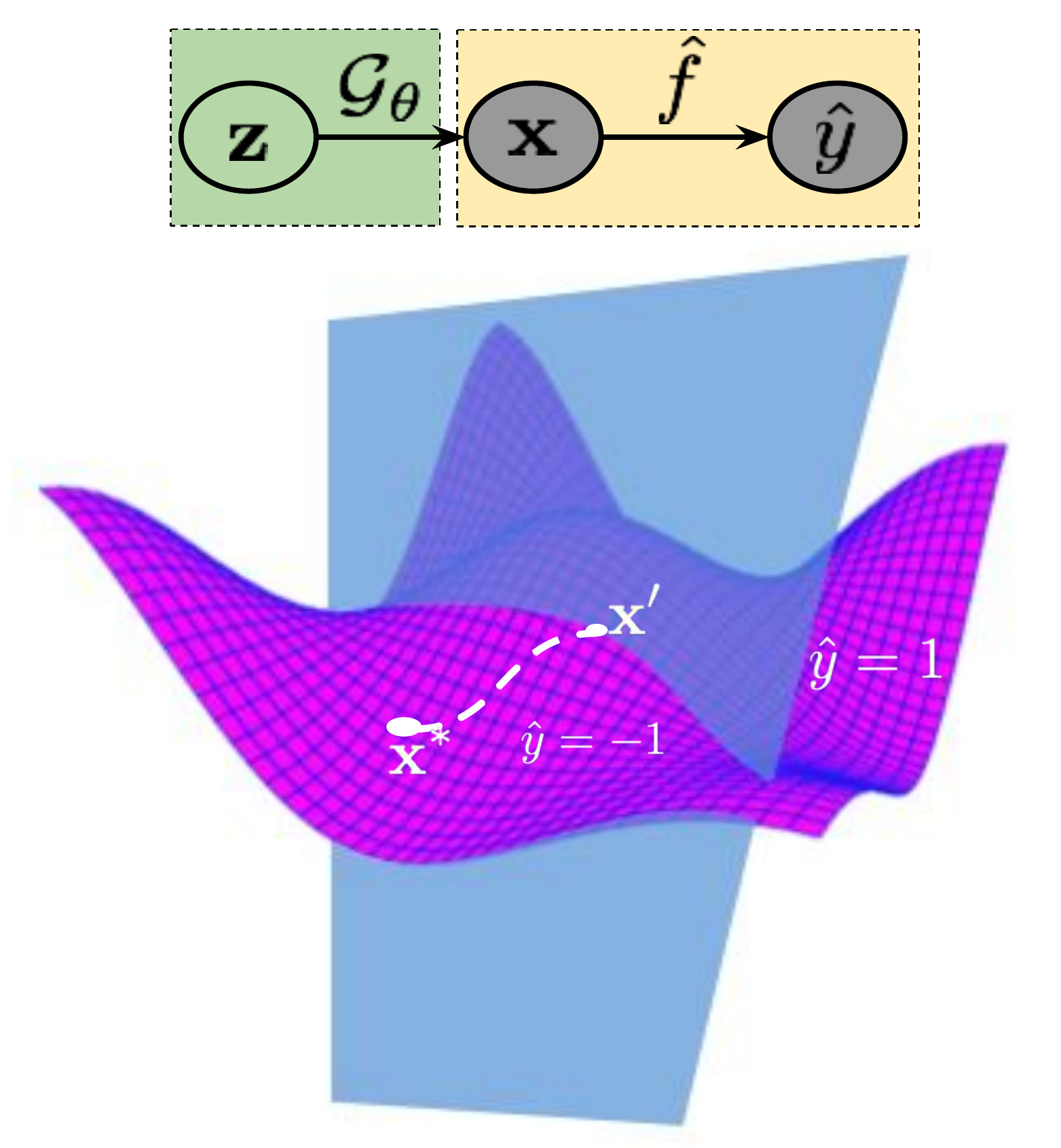}

\end{minipage}%
 \scalebox{0.92}{
\begin{minipage}{0.66\textwidth}
\centering
\begin{algorithm}[H]
\caption{\bf{\revise \\ Input:  $\bx^* \text{s.t.} \, f(\bx^*) = -1 \\ 
\mathcal{G}_{\theta}, \mathcal{F}_{\psi}, f, \lambda>0, \eta, \tau_{max}> 0, tt=0$}}
\label{alg:algo1}
\begin{algorithmic}[1]
\STATE{Initialize $\bz \leftarrow \mathcal{F}_{\psi}(\bx^*)$}
\WHILE{$f(\mathcal{G}_{\theta}(\mathbf{z})) \neq 1$ or $tt < \tau_{max}$}
\STATE{$\bz \leftarrow \bz - \eta \nabla_{\bz}(\ell (\hat{f}(\mathcal{G}_{\theta}(\mathbf{z})),1) + \lambda c(\mathbf{x}^*, \mathcal{G}_{\theta}(\mathbf{z})))$}
 \STATE{$tt \leftarrow tt + 1$}
 \ENDWHILE
 \STATE{$\bx' \leftarrow \mathcal{G}_{\theta} (\bz)$}
 \IF{$f(\mathcal{G}_{\theta}(\mathbf{z})) == 1$}
 \STATE{Return $\{(d_i, \mathbf{x}_i^* - \mathbf{x}_i') \forall i \in [d]
 \text{s. t. }  abs(\mathbf{x}_i^* - \mathbf{x}_i') >0 \}$}
 \ELSE
 \STATE{Return NULL} 
 \ENDIF
\end{algorithmic}
\end{algorithm}
\end{minipage}
}
\caption{(Top left inset) Generative model for individual recourse in supervised classification. (Bottom left inset) Illustration of procedure to obtain recourse using the data manifold in a classifcation setting. $f$ is the decision making system. The magenta surface denotes the data manifold approximated using a generative model. $\mathbf{x}^*$ is a sample with an undesirable outcome (-1) and $\mathbf{x}'$ is the sample obtained by using Algorithm~\ref{alg:algo1} (right inset) to obtain recourse using Equation~\ref{eq:classification_recourse_g}.}
\label{fig:classification_recourse}
\stepcounter{figure}%
\end{figure}

For the same set of samples used for training the decision making system, a generative model $\mathcal{G}_{\theta}$ is trained as described above. In most of our experiments we employ the decoder of VAEs as our generative model of choice. The generator thus allows us to sample $\mathbf{x} \sim \mathcal{G}_{\theta}(\mathbf{z})$ from the data distribution. Approximating the constraint as regularization, the cost function in Equation~\ref{eq:classification_recourse} can be modified as:
\begin{equation}\label{eq:classification_recourse_g}
 \mathbf{x}' = \argmin_{\mathbf{z} \sim \mathcal{G}_{\theta}(\mathbf{z})}{ \min_{\lambda}   \ell (\hat{f}(\mathcal{G}_{\theta}(\mathbf{z})),1) + \lambda c(\mathbf{x}^*, \mathcal{G}_{\theta}(\mathbf{z}))}
\end{equation}
where $\lambda>0$ determines the trade-off between the closeness of the generated recourse sample to the original sample and its corresponding target label. Intuitively, we would like the smallest possible regularization that would allow us to "cross" the decision boundary and generate a sample close to the original sample albeit with a better outcome. In our proposed algorithm, we cross-validate over $\lambda$. Figure~\ref{fig:classification_recourse} shows the graphical model corresponding to our decision making system (shaded yellow). We augment the decision making system using the generative model (shaded in green) in order to provide recourse in a classification setting by optimizing Equation~\ref{eq:classification_recourse_g}. A conceptual illustration of the desired effect is demonstrated in Figure~\ref{fig:classification_recourse}.

Our algorithm to obtain recourse proceeds as follows. We first obtain the latent encoding of our sample $\mathbf{x}^*$ using the encoder $\mathbf{z}_{0} \leftarrow \mathcal{F}_{\psi}(\mathbf{x}^*)$. For a fixed $\lambda >0$, we take gradient steps in the latent space of the generator starting from $\mathbf{z}_{0}$ so as to minimize Equation~\ref{eq:classification_recourse_g}. We take targeted gradient steps until we cross the decision boundary such that $f(\mathcal{G}_{\theta}(\mathbf{z})) = 1$. The sample $\mathbf{x}' \leftarrow \mathcal{G}_{\theta}(\mathbf{z})$ thus obtained is used to generate the appropriate recourse tuple. This procedure is summarized in Algorithm~\ref{alg:algo1}, called REVISE.

\subsection{Recourse in Causal Models}
Classification based decision making systems are limited in that they do not encode causal relationships between variables while potentially learning spurious correlations~\citep{caruana2015intelligible}. This has exposed the importance of learning and deploying causal models in practical decision making systems. In a causal decision making system, the main goal is to evaluate outcomes under different ``treatments'' and use interventions corresponding to the treatment that improves the outcome. Learning such models is challenging (without randomized control trials) due to the lack of data where counterfactuals are observed. Causal effect estimation from observational data is further compounded by (unobserved) \emph{confounders}, that affect the treatment provided as well as the outcome (for example, wealthier patients might be able afford more expensive medication). Much of the recent progress in developing such decision making systems attempt to (approximately) learn in the presence of hidden confounders~\citep{louizos2017causal,madras2018fairness} by estimating these confounders. The main assumption made is that hidden confounders can be reasonably estimated via latent variable models leveraging (approximate) learning algorithms.

\begin{figure}[!htp]
\centering
\begin{minipage}{0.5\textwidth}
\centering
\subcaptionbox{}{\includegraphics[scale=0.30]{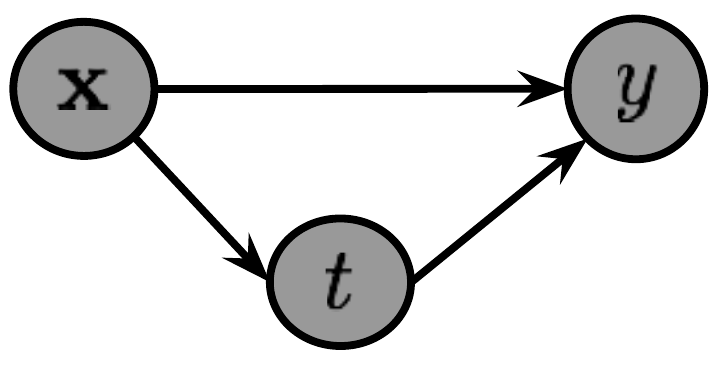}}
\end{minipage}%
\begin{minipage}{0.5\textwidth}
\centering
\subcaptionbox{}{\includegraphics[scale=0.30]{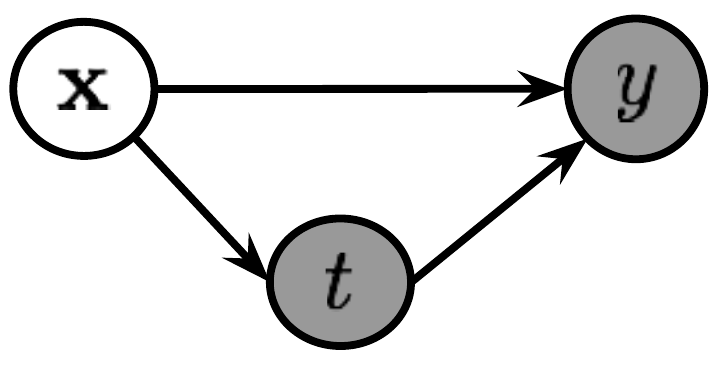}}
\end{minipage}
\begin{minipage}{0.5\textwidth}
\centering
\subcaptionbox{}{\includegraphics[scale=0.33]{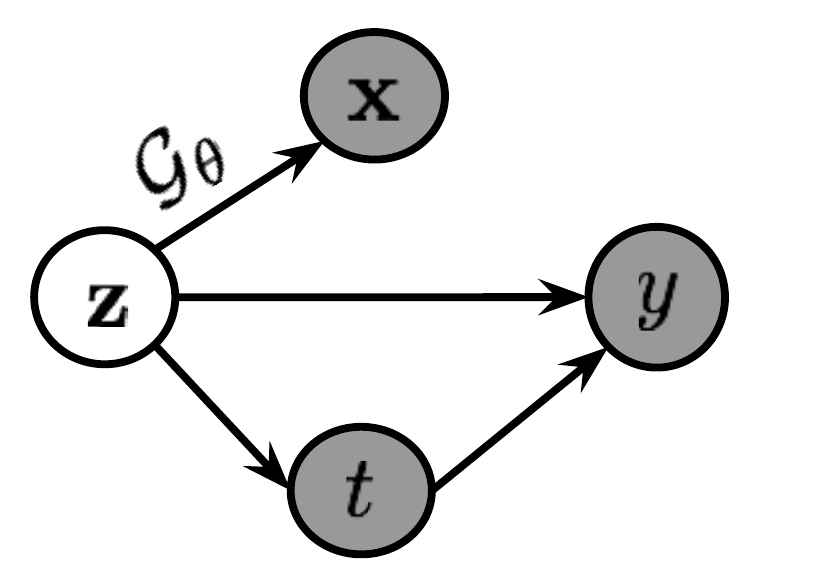}}
\end{minipage}%
\begin{minipage}{0.5\textwidth}
\centering
\subcaptionbox{}{\includegraphics[scale=0.33]{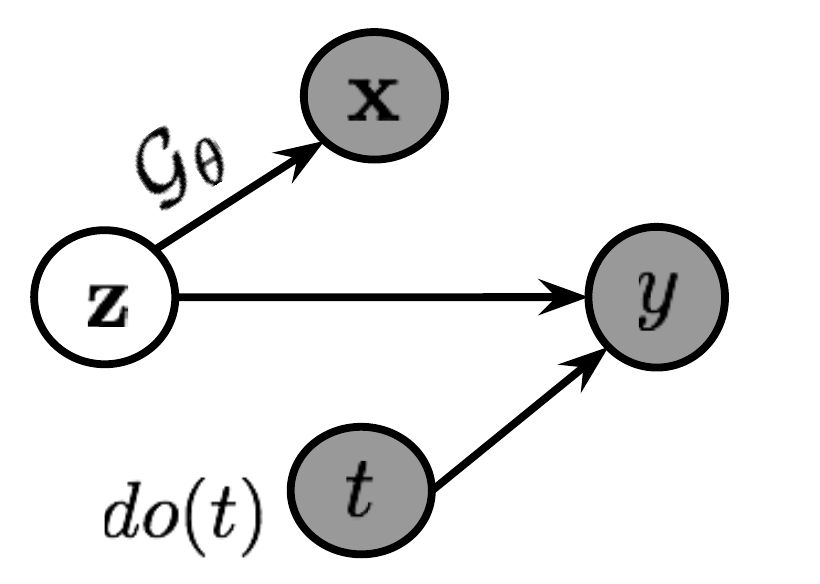}}
\end{minipage}
\begin{minipage}{0.5\textwidth}
\centering
\subcaptionbox{}{\includegraphics[scale=0.33]{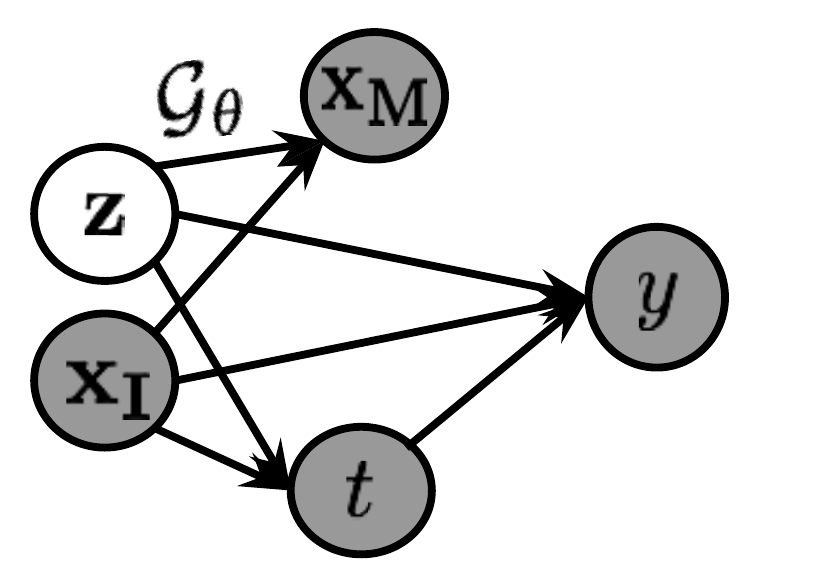}}
\end{minipage}%
\begin{minipage}{0.5\textwidth}
\centering
\subcaptionbox{}{\includegraphics[scale=0.33]{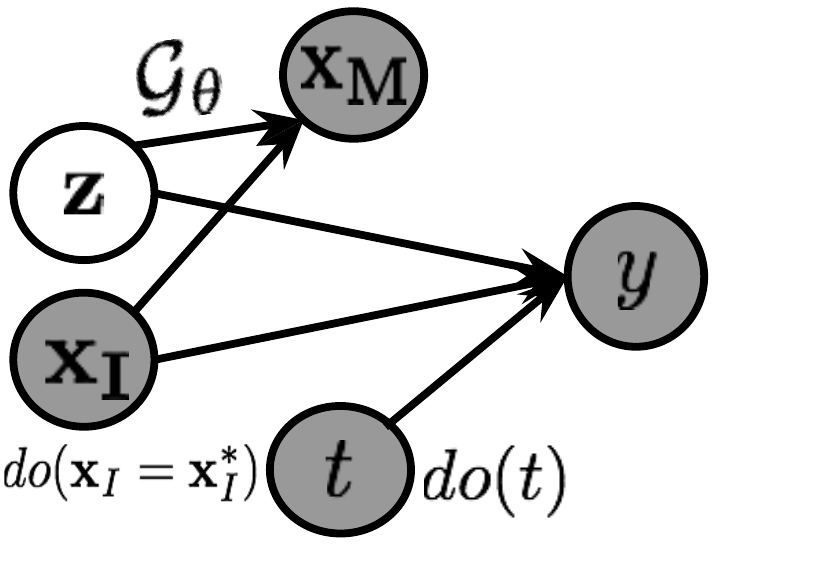}}
\end{minipage}
\caption{(a) Graphical model for a decision making system when all confounders are observed. $\bx$ is the set of observed attributes (including confounding variables) that affect the binary ``treatment'' $t$ and the outcome of interest $y$ (b) Graphical model when all confounders are unobserved. (c) Decision making system for which recourse is learned in this work. (d) Recourse under intervention (e, f) Recourse with immutable variables with appropriate intervention.}
\label{fig:caounterfactual_dms}
\end{figure}

Figure~\ref{fig:caounterfactual_dms} (a) shows a simple graphical model corresponding to such a decision making system. $\bx$ is the set of observed attributes (including confounding variables) that affect the (binary) ``treatment'' $t \in \{0, 1\}$ and the outcome $y \in \{-1, -1\}$. Figure~\ref{fig:caounterfactual_dms} (b) shows a corresponding model where confounders are not observed. The causal effect of the treatment is usually determined by ``intervening'' on the treatment variable (i.e. clamping it to a fixed value irrespective of the realizations of its parents in the associated graph) and studying the outcomes. Such interventions are codified as $do(.)$ operations in causal calculus~\citep{pearl2009causality}. When confounders are not observed as in Figure~\ref{fig:caounterfactual_dms}(b), it is difficult and in some cases impossible to identify the causal effect of the treatment.

While methods that approximately estimate hidden confounders are an empirical improvement over classification systems~\citep{louizos2017causal}, a myriad of issues ranging from mis-specification of the underlying causal model, approximations used for tractability of the latent variable estimation, and selection bias in the data can cause causal models to be less than perfect. Also, while more accurate, outcomes can be still be undesirable for many individuals scrutinized under such systems (providing treatment still does not improve outcome even though on average, treatments are effective). Thus, provision of recourse is still a necessity.

We demonstrate how the proposed algorithm ~\revise  can be modified to generate recourse for a causal decision making system. In particular, we focus on the model presented in Figure~\ref{fig:caounterfactual_dms}(c) and without loss of generality restrict to a binary outcome $\{-1, 1\}$ where $1$ is the desired outcome and $-1$ is an undesirable outcome. The key is to draw an analogy between estimating hidden confounders using the latent variable learning framework and the technique proposed here, that leverages the latent manifold to learn a minimal set of changes that improve outcomes. Specifically, we demonstrate how to naturally traverse the space of hidden confounders so as to improve the outcome $y$ (potentially under specific interventions of the treatment variable $do(t)$). In this decision making system $\mathcal{G}_{\theta}$ learns the relationship between the attributes $\bx$ and the associated hidden confounder $\bz$. Consider a data point with attributes $\bx^*$ with an undesirable outcome ($y = -1$). We obtain recourse by solving Equation~\ref{eq:counter_loss} under this causal model.
\begin{equation}\label{eq:counter_loss}
 \mathbf{x}' = \argmin_{\mathbf{z} \sim \mathcal{G}_{\theta}(\mathbf{z})}{ \min_{\lambda}   \ell (\log{y}_{do(t)},1) + \lambda c(\mathbf{x}^*, \mathcal{G}_{\theta}(\mathbf{z}))}
\end{equation}
where $\ell(.)$ is the cross-entropy function. That is, we find the minimal change in the attributes such that the causal outcome is improved. Note that this is indirectly done by re-estimating the hidden confounders that in turn affect the treatment and outcome while still being close to the original point (measured by $c(\bx^*, \bx)$). However, in practice, we would like to estimate the recourse under different interventions of the treatment $t$. We denote this outcome as $y_{do(t)}$. Normally $t$ is a binary variable (0= not treated or 1=treated).  This intervention corresponds to the causal graph shown in Figure~\ref{fig:caounterfactual_dms}(d). Of particular importance are cases where the outcome does not improve even when the treatment variable is 1, especially if the treatment on average is known to improve outcomes. This, in practice, corresponds to first intervening on the treatment variable $do(t=1)$ and providing recourse after intervention. The inference network used to estimate the posterior $p(\bz | \bx, t, y)$ is not shown for space constraints. The analogous algorithm to recommend recourse can be easily obtained by replacing the cost function of \revise with Equation~\ref{eq:counter_loss}. 

%% file: experiments.tex
\section{Experiments}\label{sec:experiments}
First, we evaluate Algorithm~\ref{alg:algo1} when the decision making system is a supervised classification black-box.
\input{recourse_classification.tex}\label{sec:classification_defaultcredit}
\input{recourse_comparison.tex}\label{sec:counterfactual}
\input{attribute_confounding.tex}\label{sec:attribute_confounding}

%% file: recourse_classification.tex
\subsection{Recourse for Classification Systems}
We provide recourse on the UCI defaultCredit~\citep{yeh2009comparisons} dataset. The goal is to improve outcomes for clients who are expected to default on credit in the next month according to a classification based decision making system. 
We preprocessed the data as closely as possible to the baseline (see Appendix~\ref{app:defaultCredit}) for details. This dataset has highly heterogeneous attributes and have to be handled accordingly to generate a viable recourse. We train an HI-VAE~\citep{nazabal2018handling} as our generative model to handle heterogeneity of attributes. The architecture and model settings for training the HI-VAE are in the Appendix. Next we train a linear softmax classifier with different levels of $\ell_1$-regularization, for which a comparison with~\citep{ustun2018actionable} is possible as well as non-linear classifiers (dense deep neural networks) to demonstrate utility of our model to a larger class of classification systems. Qualitative results are shown in Table~\ref{tab:result_default_credit_qualitative}. Quantitative results are relegated to the appendix. As can be seen from these results,~\citet{ustun2018actionable} propose recourses that are very sparse but very large in specific attributes (see `Most Recent Payment Amount for both samples'). This makes such recourses unrealistic in practice. However notice that for REVISE (MLP), "Max Bill Amount Over Last 6 Months" is less than the "Most Recent Bill Amount" suggesting a contradiction that neither of the baselines address. This contradiction can be solved by optimizing the counterfactual directly in the data domain (and applying constraints on $\bx$) with manifold constraints similar to that implemented in this work albeit as a regularization (see an alternative in~\citet{dhurandhar2018explanations}).
\begin{table*}[t]
\caption{Sample Recourse from \revise for the UCI defaultCredit dataset for a (linear) softmax classifier, (non-linear) MLP and~\citet{ustun2018actionable} for one example. `Original' features correspond to clients defaulting, recourses correspond to preventing default under different methods.`-'  implies no change was recommended for that attribute. Features not listed in the table are not recommended for recourse by any baseline. More results in Appendix~\ref{app:defaultCredit}.}
 \label{tab:result_default_credit_qualitative}
 \begin{center}
  \begin{small}
\resizebox{\textwidth}{!}{%
\begin{tabular}{lllll}
\toprule
Attribute & original &  \revise (Linear) & \revise (MLP) & Ustun et. al. `18 (Linear) \\
\midrule
Max Bill Amount Over Last 6 Months &   2240.0 &  3461.2947 & 1548.9572 & -\\
Max Payment Amount Over Last 6 Months &    110.0 &   100.3251 & 17.0988 & -\\
Months With High Spending Over Last 6 Months &      6.0 &     0.0547 & 1.9147 & -\\
Most Recent Bill Amount &   2050.0 &  1768.1843 & 2059.7888 & -\\
Most Recent Payment Amount &     80.0 &    28.2974 & 0.0 & 6010.0 \\
Total Overdue Counts &      1.0 &     1.7552 & 0.5058 & -\\
Total Months Overdue &     12.0 &       1.05 & 0.4 & -\\
Others (Marital Status) & 0.0 & - & - & 1 \\
\bottomrule
\end{tabular}}
 \end{small}
 \end{center}
\end{table*}

%% file: recourse_comparison.tex
\subsection{Recourse in Causal Models}
The goal of this evaluation is to demonstrate how recourse can be suggested in causal models. Our method is the first to the best of our knowledge to propose recourse in such settings. We evaluate \revise with the modified cost function on the decision making system provided in Figure~\ref{fig:caounterfactual_dms}(c). We evaluate the sparsity of recourse, distance in latent space as well as input space of recourses obtained using \revise. We evaluate these factors when the counterfactual models are trained on data with different biases. These biases are not simple side-effects of imbalance in labeled data as is commonly studied in classification settings. In this case, the bias is a true reflection how reliable the treatment effect estimation can be. That is we compare the case of training the model on randomized treatment assignments versus the more common observational setting. We demonstrate results on a dataset where we can simulate both cases.

\subsubsection{Handling Immutable Variables}
Immutable Variables are attributes that should remain unchanged as part of suggesting recourse, like gender, age, ethnicity. We propose to handle immutable variables by learning the observed attributes conditioned on immutable variables. Let $\bI$ index the set of immutable attributes. The set of variables allowed to change in order to suggest recourse is denoted by $\bx_{\bM}$ where $\bM = [d] \setminus \bI$. We propose to modify the existing causal decision making systems that currently don't allow for handling immutable variables easily, to instead learn a conditional causal decision making system. The corresponding graphical representation of such a decision system is shown in Figure~\ref{fig:caounterfactual_dms}(e). Note that immutable variables can be confounding variables. To recommend a recourse without allowing immutable variables to change, we fix the attributes $\bx_{I}$ to be the original attributes $\bx^*_{I}$ corresponding to the data point $\bx^*$ by conditioning on the immutable variables $\bx_{I} = \bx_{I}^*$. Note that this also allows for comparison between recourses under the causal setting "if the race had been different, what would recourse looked like?". However, discussion of such ``counterfactual'' settings is relegated to future work. For this decision making system we modify an existing causal latent variable model called CEVAE~\citep{louizos2017causal} to handle immutable variables.
\begin{table}[!htp]
 \caption{Recourse summary for conditional CEVAE on data with and without hidden confounding. The original outcome is mortality within 1 year and the recourse prevents mortality under treatment.}
 \label{tab:recourse_counterfactual}
\centering
\begin{tabular}{p{1.3cm}p{2.3cm}p{2.3cm}p{2cm}p{2cm}} 
\toprule 
Method &  Data-confounded & Median \# changes &  (Mean) $\Delta_x$ &  (Mean) $\Delta_z$ \\
\midrule
\revise & TWINS-no& 6 (max=121) &     3.1055	 &        0.0245 \\
\revise & TWINS-yes & 5 (max=121) &       2.9440 &        0.0367 \\ 
\bottomrule
\end{tabular}
\end{table}

Evaluating recourse on causal models is difficult due to the lack of counterfactual information (as we would like to intervene with treatment $do(t=0)$). In light of this, only simulated and/or limited datasets with counterfactual information can be used for such evaluation. One such dataset is the TWINS dataset provided by~\citet{louizos2017causal} which contains details of twins (of the same sex in each pair), lighter than $2\, kgs$ born between 1989 to 1991. The features used in this dataset, specifically risk factors associated with specific conditions may not always be amenable to recourse. The results are nonetheless can be useful for our demonstrations. The treatment assignment is whether a twin is heavier ($t=1$ if heavier and $0$ otherwise). The attributes are details of parents' risk of conditions and history (see Appendix~\ref{app:twins} for details). One of each pair of twins can be included for training and the counterfactual outcome is the outcome of the other twin (not included in training the model). The dataset has $~11984$ pairs of twins. The mortality outcome (which is the outcome of interest) is better for heavier twins by $\sim2.5\%$. However, the mortality rate among heavier twins is still as high as 16.4\% and suggests that even under an accurately learned model, improving outcomes for parents by suggesting recourses is beneficial. We consider two settings for evaluation- i) randomized control trial and ii) the case of hidden confounding. In the first case, treatment assignments are chosen at random while in the latter by using the number of gestation weeks as a confounding feature following the procedure described in~\citet{louizos2017causal}. In each case, recourses are summarized on the counterfactual data.

We selected `sex of child' and `birth month' as the immutable variables. Other attributes correspond to parental history and risk or propensity to conditions. While the recourses in this case are not practical (by virtue of the dataset itself), they are illustrative of how confounding can change the quality of recourse under the same class of systems. Once a conditional CEVAE is trained, recourse is obtained with and without hidden confounding by fixing these attribute to observed values in the test set. Table~\ref{tab:recourse_counterfactual} provides a summary of the recourse in terms of the sparsity of the recourse (mean number of attributes changed) as well as the latent space  and input distances for the TWINS data trained with and without confounding using \revise. Qualitative samples are provided in Figure~\ref{tab:recourse_qualitative}. Qualitatively, the amount of confounding significantly changes the nature of recourse provided to a patient. This suggests that even causal decision making systems themselves are fragile and should be cautiously deployed in practice. Lowering specific risk factors is the main set of recourses provided by this model.
\begin{table}[!htp]
\caption{Recourse on TWINS data on models trained with and without confounding under intervention $do(t=1)$. The original features correspond to mortality within 1 year and the recourse prevents mortality under treatment.`-' indicates no change recommended.}
\label{tab:recourse_qualitative}
\centering
\scalebox{0.8}{%
\begin{tabular}{cccc}
\toprule
feature name &  original &  recourse (no confounding) & recourse (confounding) \\
\midrule
risk factor Hvdramnios (0=no risk) &       1.0 &       0.0 & 0.0 \\
risk factor, Incompetent cervix (0=no risk) &       1.0 &       0.0 & 0.0 \\
total number of births before twins &       8.0 &   -    & 1.0 \\
Other Medical Risk Factors (0=no risk) &       1.0 &       0.0  & 0.0 \\
risk factor, Diabetes (0=no risk) &       0.0 &       1.0  & - \\
\bottomrule
\end{tabular}}
\end{table}

%% file: attribute_confounding.tex
\subsection{Recourse under Attribute Confounding}
Our recourse framework provides diagnostic capabilities to compare classifiers learned under biased data. For instance, a classifier trained to determine the best medical intervention may be relying on attributes like gender to determine best treatment for a (clinically known) gender neutral condition. Note that this is strictly different from relying on an immutable variable to make decisions. For instance, bias can be introduced because of imbalance in the data (not associated with an immutable variable) and therefore may not be fixed by just conditioning on bias inducing attributes. In many cases, it is not clear which attributes are confounded with the outcome. It is desirable to identify as well as monitor such behavior. We study this case more as a problem of changing from one label to another as opposed to transitioning from a bad outcome to an improved outcome. We do this experiment using image data. This evaluation provides evidence that our model can be also be used to generate counterfactual explanations.
\begin{figure}[t]
\setlength{\floatsep}{2pt plus 1.0pt minus 2.0pt}
  \includegraphics[width=\textwidth]{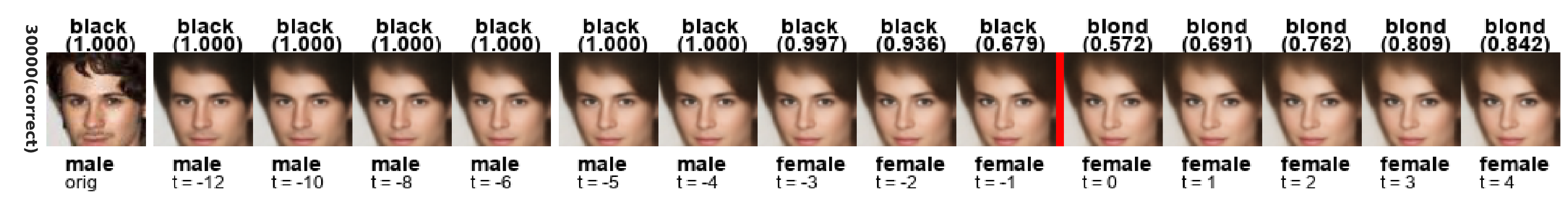}
      \vskip -8px
      \includegraphics[width=\textwidth]{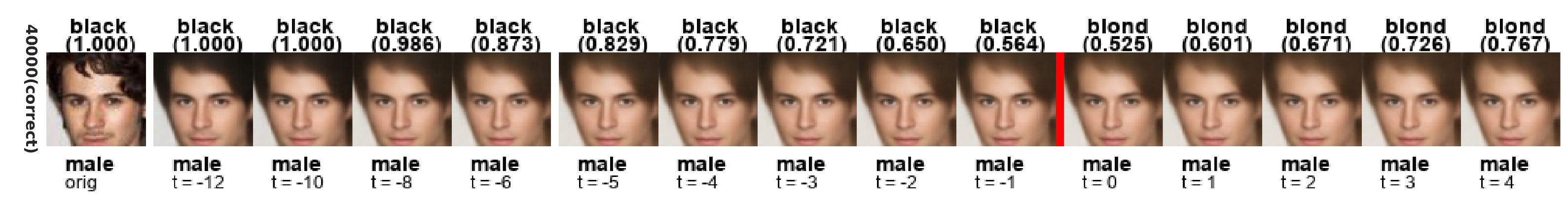}
      \medskip
     \includegraphics[width=\textwidth]{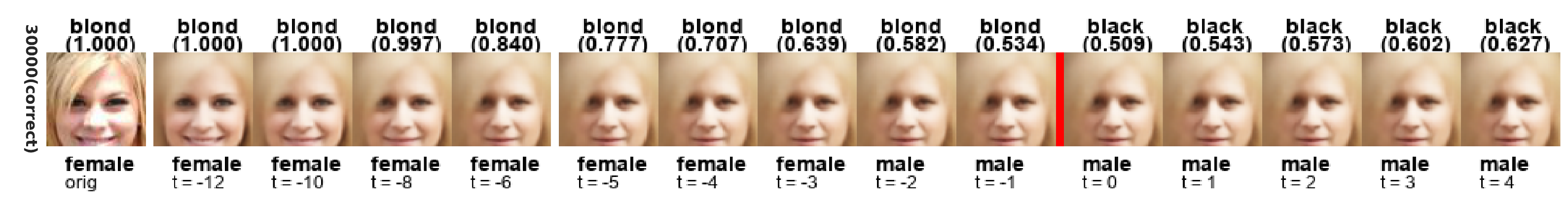}
     \vskip -12px
     \includegraphics[width=\textwidth]{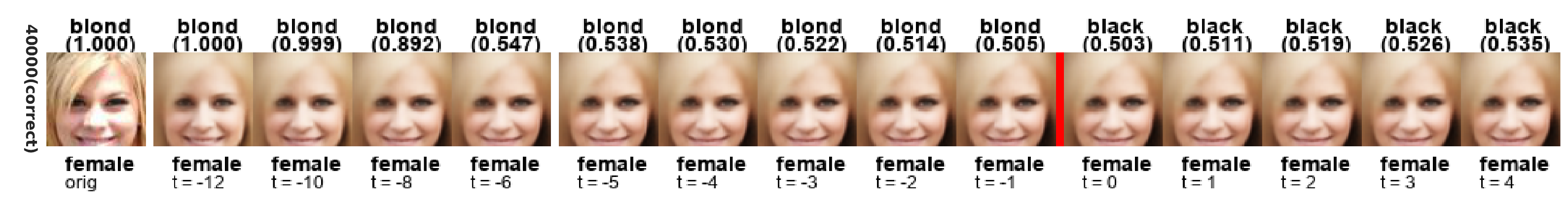}
  \caption{Two samples for classifiers $f_2$ (first sub row) and $f_1$ (second sub row) are shown. The leftmost image is the original figure, followed by its reconstruction from the VAE. Intermediate reconstructions are shown as Algorithm~\ref{alg:algo1} progresses toward crossing the decision boundary (with $\lambda=0.01$). The red bar indicates change in hair color label indicated at the top of each image along with the confidence of prediction. $t$ at the bottom of each image corresponds to the iterations in Algorithm~\ref{alg:algo1}, shifted by $T$, where $T$ is the iteration where label flips. For both samples, biased classifier $f_2$ shows demonstrable changes in gender specific features ($1^{st}$ and $3^{rd}$ rows) while crossing the decision boundary. At the bottom, we show labels as predicted by a gender classifier for reference.}
  \label{fig:confounding_faces}
  \vskip -0.1in
\end{figure}

Using this example, we highlight significant shortcomings of automatic gender recognition systems that have already been rightfully criticized for potential harms on trans and non-binary individuals~\citep{Keyes:2018:MMT:3290265.3274357}. Specifically, we hope to demonstrate how simple biases induced in the dataset on auxiliary attributes (like hair color) can affect complex automated decision making systems even when they have non--trivially high performance and therefore the nature of recourse learned. This suggests their biases can be harmful when deployed in practice. We evaluate results of a (binary) gender recognition deep neural network classifier trained on celebA~\citep{liu2015faceattributes} face images. We would like to qualitatively demonstrate how transitioning between hair colors can identify whether the target black-box is biased by this attribute. From the task description, it is clear that hair color in and of itself cannot be factor in determining gender unless the dataset confounds gender with hair color. 

Two additional black-box classifiers are trained predict hair color (black or blond). We diagnose qualitatively whether or not the gender classification black-box is confounded with this attribute. To do so, we use the same recourse technique proposed in this work. However, we look at all class label transitions since there is no worse or better outcome in this case. We use Algorithm~\ref{alg:algo1} to change faces with black hair to faces with blond hair and vice–versa for both black-boxes and show how bias induced in the hair classification in turn affects the gender classification. The two black-box classifiers (ResNet models) $f_1$ and $f_2$ are trained to detect the hair color using different subsets of celebA training samples. $f_1$ is trained on the standard training split provided by~\citep{liu2015faceattributes} (this split already is biased such that most blond haired persons are annotated as female) whereas $f_2$ is trained with a dataset such that all black haired samples are male while blond haired samples are female. That is, the second black-box is trained to have a significantly worse data induced bias. The hypothesis is, clearly the black-box $f_2$, trained on biased data must learn there is a high correlation between hair color and gender. Therefore, if we were to change a black-haired sample to a blond haired sample (as a recourse and vice--versa), the gender classifier will perceive changed gender attributes more often when seeing images recoursed by this classifier than a classifier not explicitly trained on biased data. 

Additionally, a VAE that generates face images is trained on the celebA training split without any supervision (i.e. no access to hair color label or gender). The architecture of the VAE used is provided in Appendix~\ref{app:celebA}. Using the decoder of this VAE as our generator $\mathcal{G}_{\theta}(\mathbf{z})$, we run Algorithm~\ref{alg:algo1} on samples with both black hair and blond hair with the target outcome set to the complementary class. We also visualize all intermediate samples $\mathcal{G}_{\theta}(\mathbf{z}_{tt})$ (for iteration $tt$) generated in each iteration of the Algorithm. Instead of visualizing the tuples returned by Algorithm~\ref{alg:algo1}, we visualize the whole image.

Figure~\ref{fig:confounding_faces} shows two examples (from the held-out set) for both black-boxes along with the intermediate samples generated by our algorithm. One sample with black-hair ``recoursed" to have blond hair and the other sample with blond hair recoursed to have black hair. The red line marks the decision boundary or the point where the most likely label of the image flips to the complementary class. As can be seen from the figure, examples where the black-box was trained on biased ($f_2$) data for hair color classification changes gender specific attributes of the sample as it crosses the decision boundary whereas the black-box trained on unbiased data does not\footnote{Qualitative figures were chosen based on the confidence of the prediction and that of the reconstructed image}. 
Table~\ref{tab:class_details} demonstrates clearly that the fraction of times gender flipped for the biased classifier is significantly higher than of the unbiased classifier. The implication of this for our main task of providing recourse is that a decision making system trained on biased data will provide a recourse that will more often prefer a specific gender under the recoursed label and the qualitative bias therein specifically highlights dangers of relying on binary gender recognition systems to identify what recourse may look like. Thus the recourse generating algorithm can be used to identify whether such a bias is incorporated in the classifiers.

%% file: conclusion.tex
\section{Conclusions}
This work addresses the problem of algorithmic recourse. Given an individual that faces an undesirable outcome under a decision making system, we propose a mechanism to suggest a recourse, or a minimal set of changes to improve their outcome. We provide a gradient based algorithm that allows to sample from the latent space of the generative model to find the smallest set of changes that would improve outcomes. The proposed algorithm provides recourse for classification and causal decision making systems. We justify why counterfactual models should also be able to provide recourse. To handle immutable variables, we propose and train conditional variants of causal decision making systems. The proposed recourse algorithm can highlight whether a decision making algorithm is systematically confounding specific attributes. Our method thus has added diagnostic capabilities. We hope to highlight the importance of providing such algorithms alongside (potentially fair) decision making systems so that individuals can always improve their outcomes.

%% file: appendix.tex
\newpage
\section{Appendix}

All experimental evaluations follow three main steps:\begin{inparaenum}\item[i)] Train Generative models to approximate the data distribution and/or manifold \item[ii)] Train black-box models that will be candidate models for which recourse will be provided. \item[iii)] Evaluation of the recourse algorithm proposed in Algorithm~\ref{alg:algo1}.\end{inparaenum}

\subsection{Evaluation: Recourse on UCI default Credit}\label{app:defaultCredit}

\begin{table}[!htp]
\centering
\begin{tabularx}{\columnwidth}{c|X|X|p{1.5cm}}
\hline
Method-($\lambda$)  &  Mean $\Delta z$ &  Mean $c(.)$ &  (Median) \# Changes \\
\midrule
\revise-1e-05 &  {\bf{0.006727}}	& {\bf{3607.4}}	& 9.0 \\
\revise-0.001	&  0.006810 & 3607.9	 &9.0\\
\revise-0.1	& 	0.009128 & 3607.6	& 9.0 \\
\revise-10.0	&  0.476822	& 3607.5	& 9.0\\
Ustun et. al. '18 &    NA & NA & {\bf{2.0}} \\
\bottomrule
\end{tabularx}
\caption{Recourse comparison for linear softmax classifier with $\ell_1$-regularization parameter set to $1e^{-5}$. Accuracy=79.37\%. Original outcome is a client defaulting on credit and the recoursed outcome is prevention of default in the next month.}
\label{tab:recourse_linear}
\end{table}

\begin{table*}[t]
\caption{Additional Qualitative Results for recourse on UCI data.}
 \label{tab:result_default_credit_qualitative2}
 \begin{center}
  \begin{small}
\resizebox{\textwidth}{!}{%
\begin{tabular}{llllp{3cm}}
\toprule
Attribute & original &  \revise (Linear) & \revise (MLP) & Ustun et. al. `18 (Linear) \\
\midrule
Education Level                              &  University &   Graduate & - &  - \\
 Max Bill Amount Over Last 6 Months           &      4000.0 &  3770.5771 & 3028.146& -\\
 Max Payment Amount Over Last 6 Months        &       370.0 &  241.5032 & 639.1942 & -\\
 Months With Low Spending Over Last 6 Months  &         0.0 & - & 0.0745 & -\\
 Months With High Spending Over Last 6 Months &         6.0 & 0.0 & 3.0379& -\\
 Most Recent Bill Amount                      &      3780.0 &  3122.0967 & 4995.4946 & -\\
 Most Recent Payment Amount                   &         0.0 &    28.0093 & 6210.4756 & 5760.0\\
 Total Overdue Counts                         &         1.0 &     1.0941 & 0.7319& - \\
 Total Months Overdue                         &        12.0 &     1.2939 & 0.0 & -\\
 Others (Marital Status)                                 &           0  &   - & - & 1 \\
\bottomrule

\end{tabular}}
 \end{small}
 \end{center}
\end{table*}

Since we explicitly optimize for $c(.)$ while~\citet{ustun2018actionable} optimize for a separate user defined cost function, in order to ensure a fair comparison, we only compare to their proposed algorithm w.r.t. number of attributes changed along with demonstrating qualitatively the difference between recourse generated by the proposed algorithm and this baseline. Since their algorithm generates multiple flipsets, we took the sparsest set among their solutions as a baseline.
\begin{table}[t]
\centering
\begin{tabularx}{\columnwidth}{c|X|X|X}
\toprule
Method-($\lambda$)  &  Mean $\Delta z$ &  Mean $c(.)$ &  (Median) \# Changes (max=20) \\
\midrule
\revise-1e-05 &  0.000997	& 3672.8	& 9.0 \\
\revise-0.001	&  {\bf{0.000931}}	& 3672.6	& 9.0\\
\revise-0.1	& 	0.003565 & 3672.7	& 9.0 \\
\revise-10.0	&  0.243121	& {\bf{3672.6}}	& 9.0\\
Ustun et. al. '18 & NA & NA & NA \\
\bottomrule
\end{tabularx}
 \caption{Recourse comparison for nonlinear multi-layer neural network classifier with $\ell_1$-regularization parameter set to $1e^{-4}$. Accuracy=80.18\%. Original outcome is a client defaulting on credit and the recoursed outcome is prevention of default in the next month.}
 \label{tab:recourse_mlp}
 \end{table}

The UCI defaultCredit dataset was processed according to scripts available here: \url{https://github.com/ustunb/actionable-recourse} for comparison with the baseline, without the gender and dropping the redundant attribute 'HistoryOfOverduePayments'. Additionally, we one-hot encode the Education variables due to issues with training HI-VAE with data as processed in the baseline repository. The data was split into train (60\%), test (20\%), and validation (20\%). First an HI-VAE\footnote{\url{https://github.com/probabilistic-learning/HI-VAE}} is trained using training data without labels. Next, classifiers listed in Table~\ref{tab:defaultcred_classifier_perf} are trained on the same subset of training data. All recourse results are then evaluated on held-out test data.

Datatype settings for defaultCredit dataset to train the HI-VAE:
\begin{table}[t]
    \centering
    \begin{tabular}{p{4cm}|p{2cm}|p{1.3cm}}
    \hline
    Attribute & Type & Dimension \\
    \hline
       Marital Status  &  Categorical & 3\\
       Age  & Categorical & 4 \\
       Education & Categorical & 4\\
       Max Bill Amount Over Last 6 Months & Positive Real & 1 \\
       Max Payment Amount Over Last 6 Months& Positive Real & 1 \\
       Months With Zero Balance Over Last 6 Months& Positive Real & 1 \\
       Months With Low Spending Over Last 6 Months& Positive Real & 1 \\
       Months With High Spending Over Last 6 Months& Positive Real & 1 \\
       Most Recent Bill Amount& Positive Real & 1 \\
       MostR ecent Payment Amount& Positive Real & 1 \\
       Total Overdue Counts& Positive Real & 1 \\
       Total Months Overdu& Positive Real & 1 \\
       \hline
    \end{tabular}
    \caption{Datatypes for processing defaultCredit dataset preprocessed according to~\citep{nazabal2018handling} to train HI-VAE}
    \label{tab:hivae}
\end{table}

\paragraph{HI-VAE settings}

\begin{table}[t]
    \centering
    \begin{tabular}{c|c}
    \hline
       Setting  & value \\
       \hline
    epochs & 1230 \\
    model name & inputDropout  \\
    batch size & 1000 \\
    dim s & 2 \\
    dim z  & 5 \\
    dim y & 5 \\
    \hline
    \end{tabular}
    \caption{Settings for training HI-VAE for the defaultCredit data without missing values to reproduce experiments in Section~\ref{sec:experiments}}
    \label{tab:defaultcred_classifier}
\end{table}

Settings for training the HI-VAE from~\citep{nazabal2018handling} are provided in Table~\ref{tab:defaultcred_classifier}

\paragraph{Decision Making Systems} Classifiers tested for recourse on defaultCredit dataset. Primarily a single layer linear classifier (Softmax) and an MLP classifier (3 hidden layers with relu activation, final layer with softmax activation) were trained with different levels of $\ell_1$- regularization:
\begin{table}[t]
    \centering
    \begin{tabular}{c|c|c}
    \hline
      Model   & $\ell_1$–regularization & Test Accuracy \\
      \hline
        Softmax  & $0.0$ & 0.7923 \\ 
        Softmax  & $10^{-5}$ & 0.7937\\
        Softmax  & $10^{-4}$ & 0.7930 \\
        Softmax  & $10^{-3}$& 0.7867\\
        Softmax  & $10^{-2}$& 0.7860 \\
        mlp  & $0.0$ &  0.7995 \\ 
        mlp  & $10^{-5}$ & 0.7987 \\
        mlp  & $10^{-4}$ & 0.8018 \\
        mlp  & $10^{-3}$& 0.7990 \\
        mlp  & $10^{-2}$& 0.786 \\
        \hline
    \end{tabular}
    \caption{Classifiers evaluated for recourse for defaultCredit data}
    \label{tab:defaultcred_classifier_perf}
\end{table}

\begin{table*}[t]
\caption{Additional recourse results for a candidate sample from the proposed recourse mechanism for the UCI defaultCredit dataset for a (linear) softmax classifier, (non-linear) MLP and~\citet{ustun2018actionable}.`-'  implies no change was recommended for that attribute by the corresponding Algorithm}
\label{tab:app1}
 \vskip 0.15in
 \begin{center}
 \begin{small}
\resizebox{\textwidth}{!}{%
\begin{tabular}{llllp{2.2cm}}
\toprule
Attribute & original &  \revise (Linear) & \revise (MLP) & Ustun et. al. `18 (Linear) \\
\midrule
Max Bill Amount Over Last 6 Months          &  760.0 & 889.0583 & 522.8529 & - \\
Max Payment Amount Over Last 6 Months       &   60.0   & 47.5637 & 66.8995 & - \\
Months With Zero Balance Over Last 6 Months &  0.0 & 0.1126 & - & - \\
Months With High Spending Over Last 6 Months &  0.0 &  0.3784 & 2.8543 & -  \\
Months With Low Spending Over Last 6 Months &  0.0 & - & 0.075 & - \\
Most Recent Bill Amount                     & 670.0 & 921.6386 & 578.4103 & - \\
Total Overdue Counts                        &  1.0 & 1.6596 &  0.5206& -\\
Total Months Overdue                        &  12.0 & 0.2819 & 0.4318  & -\\
Marital Status (Other)                      &  0&  - & - & 1\\
Most Recent Payment Amount                  & 50.0 & 11.6078 &  9.0905& 5735.0 \\
\bottomrule
\end{tabular}}
 \end{small}
 \end{center}
 \vskip -0.1in
\end{table*}


\paragraph{Additional Results}
Please see Tables~\ref{tab:app1} for additional sample results.

\subsection{Evaluation:Recourse on TWINS dataset- Comparison with Counterfactual decision making systems}\label{app:twins}
The TWINS dataset was processed using the procedure described in~\citep{louizos2017causal}. Additionally, the following features were dropped from analyses: \\
(brstate,brstate\_reg,data\_year,stoccfipb,stoccfipb\_reg,\\
birattnd,mplbir,mrace,frace,orfath,ormoth,pre4000,preterm) resulting in a total of 121 mutable variables. 

Sex of child (csex) and birth month (birmon) are used as immutable variables (13 immutable variables). 

The implementation provided here:{\url{https://github.com/AMLab-Amsterdam/CEVAE}} is modified to incorporate conditioning for immutable variables. Additional modifications include changing the outcome to binary by modifying the outcome distribution to Bernoulli. All parameters were set to default.

\subsection{Evaluation: CelebA Experiment Details}\label{app:celebA}
\paragraph{Generator Settings} For this experiment the standard train-test split provided by~\url{http://mmlab.ie.cuhk.edu.hk/projects/CelebA.html} is used. First, a VAE is trained on the training split (without attribute information) to generate face images. The VAE used is available here:\url{https://github.com/LynnHo/VAE-Tensorflow}. All settings and architectures were set to default values. Note that all faces (brown haired, black haired, as well as blond haired faces) are used.

\paragraph{Classifer Settings}
The models evaluated in Section~\ref{sec:attribute_confounding} are hair color classifers. While these are not necessarily recourse models, they have been used for demonstrating qualitative diagnostic evaluation. The architecture and code of the ResNet models used as available here:~\url{https://github.com/ry/tensorflow-resnet}. Two models with the same architectures are trained with different subsets of data. Model 1 ($f_1$) is trained such that from the original training split, only male samples with black hair and only female samples with blond hair are used (all brown hair samples are removed from the training split). The gender attribute can be obtained from the attribute meta-data provided here:~\url{http://mmlab.ie.cuhk.edu.hk/projects/CelebA.html}. Model 2 ($f_2$) is trained on the entire training split with brown hair samples removed. Note that the generator is trained on the entire training split (including faces with brown hair).
\begin{table}[h]
\centering
\begin{tabular}{p{2.2cm}ll}
\toprule
& \multicolumn{2}{c}{Target black-box label} \\
\cmidrule(r){2-3}
Attribute ($a$) Classifier     & Black Hair     & Blond Hair \\
\midrule
$g$ (orig) & FP:0.003  & FP:0.000     \\
& FN:0.002  & FN:0.018     \\
& Acc: 0.997  & Acc:0.999    \\
\midrule
$g$ (recalibrated) & FP:0.003  & FP:0.003 \\
& FN:0.018  & FN:0.018  \\
& Acc:0.989  & Acc:0.996 \\
\bottomrule
\end{tabular}
\vspace{5px}
\small
\captionof{table}{Recalibrated Gender Classifier}
\label{tab:gender_classifier}
\end{table}
       
The gender classifier trained for reference is also a ResNet model following the architecture here:~\url{https://github.com/ry/tensorflow-resnet}. This classifier is recalibrated to have equal error rates across both hair color labels and the performance is summarized in Table~\ref{tab:gender_classifier}.

\begin{table}[t]
\centering
\begin{tabular}{p{2cm} lp{2.5cm}}
\toprule
Black-box Classifier & Accuracy & Fraction of times gender switched during recourse \\
\midrule
$f_1$ & 0.9933 & 0.1704\\
$f_2$ &0.9155 & 0.4323\\
\bottomrule
\end{tabular}
\captionof{table}{Classifier performance on CelebA data trained using ResNet models and fraction of times a gender flip was detected while changing hair color.}
\label{tab:class_details}
\end{table}
